\documentclass[letterpaper]{article} 
\usepackage{aaai2027}  
\usepackage[hyphens]{url}  
\usepackage{graphicx} 
\usepackage{natbib}  
\usepackage{caption} 
\usepackage{amsmath}
\usepackage{amssymb}
\usepackage{booktabs}
\usepackage{xcolor}
\usepackage{colortbl}

\definecolor{accblue1}{HTML}{EFF6FC}
\definecolor{accblue2}{HTML}{D8E9F7}
\definecolor{accblue3}{HTML}{B8D6EF}
\definecolor{accblue4}{HTML}{86B8E0}
\definecolor{accblue5}{HTML}{4E91CF}
\definecolor{accblue6}{HTML}{1F5FA8}
\definecolor{accblue7}{HTML}{08306B}

\title{FACT: A Forensic Agent with Compiled Tool-Use Trajectories for AI-Generated Image Detection}
\author{
Jiaoyang Chen\textsuperscript{\rm 1,2,*},
Bin Hu\textsuperscript{\rm 3},
Jingyu Hu\textsuperscript{\rm 4},
Kun Zhou\textsuperscript{\rm 5},
Qin Zhang\textsuperscript{\rm 5},
Zhengzhe Liu\textsuperscript{\rm 1}\corresponding
}
\affiliations{
\textsuperscript{\rm 1}Lingnan University, Hong Kong\\
\textsuperscript{\rm 2}Harbin Institute of Technology, Shenzhen\\
\textsuperscript{\rm 3}The University of Hong Kong\\
\textsuperscript{\rm 4}The Chinese University of Hong Kong\\
\textsuperscript{\rm 5}Shenzhen University\\
\textsuperscript{*}Work done during an internship at Lingnan University, Hong Kong.\\
zhengzheliu@LN.edu.hk
}

\begin{document}
\maketitle

\begin{abstract}
AI-generated image detection is increasingly open-world: new image
generators produce highly realistic images that make visual artifacts
harder to identify. Existing detectors usually rely on a fixed set of
forensic cues, so a detector that works well for one generator family may
fail on another. We introduce \textbf{FACT}
(\textbf{F}orensic \textbf{A}gent with \textbf{C}ompiled
\textbf{T}ool-use Trajectories), which learns an image-conditioned
tool-use policy for forensic analysis. Instead of applying a fixed
detector, FACT decides which forensic tools to call, interprets the
returned evidence, and stops when sufficient evidence has been collected.
FACT follows an \emph{Evolve--Distill--Refine} pipeline: it evolves an
execution-verified forensic skill, compiles the skill into
action--observation tool-use trajectories, distills them into a compact agent, and refines the policy with cost-aware GRPO. Across two internal and four public benchmarks, FACT achieves the best performance among all compared methods, including on recent generators held out from all FACT training stages, deepfakes, and manipulated images.

\end{abstract}


\section{Introduction}
\label{sec:introduction}

\begin{figure}[t]
\centering
\IfFileExists{figures/fig1.pdf}{
    \includegraphics[width=\columnwidth]{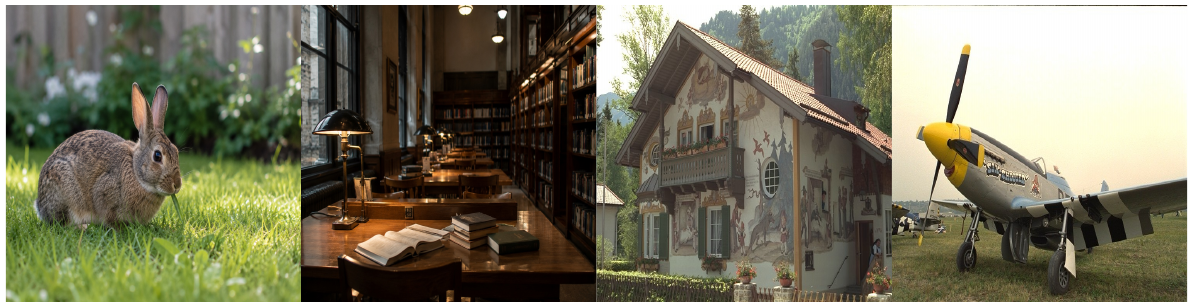}
}{
    \IfFileExists{figures/fig1.png}{
        \includegraphics[width=\columnwidth]{figures/fig1.png}
    }{
        \includegraphics[width=\columnwidth]{figures/fig1_motivation.png}
    }
}
\caption{Can you tell which images are real and which are generated by
AI?\protect\footnotemark}
\label{fig:motivation}
\end{figure}

The visual boundary between photographs and AI-generated images is
rapidly disappearing. Modern generators reproduce coherent geometry,
realistic materials, legible text, and plausible lighting, leaving few
artifacts that casual inspection can reliably identify. Figure~\ref{fig:motivation}
turns this ambiguity into a simple question: which images are synthetic?
The same ambiguity also challenges automatic detection. As image
generators become more realistic and diverse, a cue that separates real
and synthetic images for one generator family may not remain reliable
for another. Image authentication is therefore an open-world problem.

This open-world setting is difficult for a single fixed detector because
existing methods rely on different forensic priors: CNN-synthesis
artifacts~\cite{wang2020cnn}; representations from large pretrained
vision--language models~\cite{ojha2023universal}; generator-diverse
training across thousands of models in Community
Forensics~\cite{park2024community}; high-level semantic and
high-/low-frequency features in AIDE~\cite{yan2025sanity};
camera-derived photographic priors in SDAIE~\cite{zhong2025sdaie};
multimodal detection, localization, and explanation in
SIDA~\cite{huang2024sida}; and pattern-aware planning and self-reflection
in Veritas~\cite{tan2025veritas}. These cues are useful, but their
reliability varies across generator families. In our balanced evaluation
in Table~\ref{tab:detector_specialization}, Veritas, AIDE, Community
Forensics, and SDAIE each lead on different generator subsets. A fixed
detector therefore commits to one evidence profile before seeing the
image. The central question is instead: \emph{which evidence should be
acquired for this particular image?}

\footnotetext{In Fig.~\ref{fig:motivation}, the first two images are
AI-generated, while the last two are real photographs.}

\begin{table}[t]
\centering
\small
\setlength{\tabcolsep}{1.0pt}
\renewcommand{\arraystretch}{1.08}
\begin{tabular}{@{}lcccccc@{}}
\toprule
Method
& \shortstack{Nano\\Banana}
& \shortstack{Nano\\Pro}
& \shortstack{GPT-\\Image-1}
& \shortstack{SD3.5\\L}
& \shortstack{SD3\\M}
& \shortstack{Z-Image\\Turbo} \\
\midrule
VERITAS
& \cellcolor{accblue5}{\textbf{0.74}}
& \cellcolor{accblue5}{0.75}
& \cellcolor{accblue5}{\textbf{0.76}}
& \cellcolor{accblue2}{0.50}
& \cellcolor{accblue4}{0.64}
& \cellcolor{accblue1}{0.35} \\

SIDA
& \cellcolor{accblue4}{0.70}
& \cellcolor{accblue3}{0.61}
& \cellcolor{accblue2}{0.51}
& \cellcolor{accblue2}{0.53}
& \cellcolor{accblue3}{0.58}
& \cellcolor{accblue3}{0.57} \\

SDAIE
& \cellcolor{accblue4}{0.64}
& \cellcolor{accblue2}{0.54}
& \cellcolor{accblue1}{0.36}
& \cellcolor{accblue4}{\textbf{0.69}}
& \cellcolor{accblue4}{0.70}
& \cellcolor{accblue7}{\color{white}\textbf{1.00}} \\

AIDE
& \cellcolor{accblue2}{0.52}
& \cellcolor{accblue6}{\color{white}\textbf{0.84}}
& \cellcolor{accblue3}{0.55}
& \cellcolor{accblue2}{0.49}
& \cellcolor{accblue2}{0.50}
& \cellcolor{accblue7}{\color{white}0.98} \\

Comm.
& \cellcolor{accblue4}{0.67}
& \cellcolor{accblue3}{0.56}
& \cellcolor{accblue3}{0.59}
& \cellcolor{accblue2}{0.52}
& \cellcolor{accblue6}{\color{white}\textbf{0.83}}
& \cellcolor{accblue5}{0.71} \\
\bottomrule
\end{tabular}
\caption{Detector specialization across recent generators. Each column
is a balanced real-versus-synthetic subset. ``Comm.'' denotes Community
Forensics; L and M denote Large and Medium. Darker cells indicate higher
accuracy, and bold values mark the best detector in each subset.}
\label{tab:detector_specialization}
\end{table}

We address this image-conditioned evidence-acquisition problem with an
agentic formulation: the model must decide which evidence to acquire,
interpret returned observations, and stop when the evidence is
sufficient. The most recent published agentic framework, UniShield,
takes an initial step by routing each image to a predicted forgery domain
and expert type before executing the selected detector~\cite{huang2025unishield}.
However, this remains closer to one-step routing than to a full forensic
investigation. In open-world AI-generated image detection, useful
evidence can vary from image to image even within the same broad forgery
category. A forensic agent should therefore update its decision as new
observations arrive and determine whether additional evidence is still
needed.

The limitation of one-step routing suggests that a forensic agent must
do more than choose a detector once. A deployable agent must satisfy
three requirements. First, it needs \textbf{procedural knowledge}:
image-level labels do not specify how to investigate, which evidence to
acquire, or how to reconcile conflicting cues. Second, it needs
\textbf{specialization}: a general-purpose model can discover
procedures, but deployment requires a compact agent specialized for
forensic tool use and evidence-conditioned decisions. Third, it needs
\textbf{budget-aware stopping}: the agent must decide when another tool
call justifies its cost and when further evidence may become redundant
or misleading under changing image quality and disagreement among
heterogeneous forensic experts.

In this paper, we introduce \textbf{FACT}
(\textbf{F}orensic \textbf{A}gent with \textbf{C}ompiled
\textbf{T}ool-use Trajectories), an
Evolve--Distill--Refine pipeline for learning a deployable forensic
investigation policy. To acquire \textbf{procedural knowledge},
\emph{Evolve} represents forensic expertise as an explicit skill and
improves it through a propose--execute--verify loop, accepting a revision
only when rerunning the full tool interaction improves measured
behavior. To achieve \textbf{specialization}, \emph{Distill} uses the
verified skill to guide a strong teacher in the real tool environment,
compiling the procedure into action--observation trajectories and
behaviorally cloning them into a compact agent. To enable
\textbf{budget-aware stopping}, \emph{Refine} applies cost-aware GRPO
after distillation, rewarding final correctness while penalizing
unnecessary evidence acquisition. 
At deployment, FACT uses only the compact forensic agent. For each test
image, the agent decides whether to call a forensic tool, use the
returned observation, or stop with a final real/AI decision. The
skill-evolution components and the general model are used only during
training.

We evaluate FACT on two balanced internal benchmarks and four public
benchmarks. STD3K evaluates robustness under source diversity, containing
3,200 images from 21 generator/source groups and multiple real-image
sources; the final refined FACT checkpoint achieves 91.40\%. NewGen-900
evaluates generalization to nine recent generator families held out from
all FACT training stages; FACT obtains 82.44\% without
generator-specific fine-tuning.

On public benchmarks, FACT obtains 87.53\% on
Chameleon~\cite{yan2025sanity}, 88.35\% balanced accuracy on
LOKI~\cite{ye2025loki}, 98.77\% on
AIGCDetectionBenchmark~\cite{zhong2023patchcraft}, and 94.44\% on
HydraFake~\cite{tan2025veritas}. Across all six benchmarks, FACT achieves the best performance among the
methods compared under each corresponding protocol, covering
visually challenging synthetic images, recent generators held out from
all FACT training stages, and
broader forgery settings involving deepfakes and manipulated images.

Our contributions are threefold:
\begin{itemize}
    \item We formulate AI-generated image detection as an
    \emph{image-conditioned tool-use problem}. Rather than applying a
    fixed detector, the agent selectively acquires signal-level and
    model-based forensic evidence, updates its decision from returned
    observations, and stops when the evidence is sufficient.

    \item We propose an \emph{Evolve--Distill--Refine} pipeline that
    learns such a forensic agent from image-level labels. It evolves an
    executable forensic skill, compiles the skill into
    action--observation tool-use trajectories, distills them into a
    specialized policy, and further optimizes tool selection and stopping
    with cost-aware GRPO.

    \item We demonstrate strong performance and broad applicability
    across six benchmarks. FACT consistently outperforms its constituent
    tools, majority-vote ensembling, and recent detector baselines, while
    generalizing to recent generators held out from all FACT training stages
and remaining effective on
    deepfakes and manipulated images.
\end{itemize}

\section{Related Work}

\paragraph{AI image generation.}
Image synthesis has evolved from GAN-based generation
\cite{goodfellow2014gan,karras2019stylegan}, through diffusion and
latent-diffusion models \cite{ho2020ddpm,rombach2022ldm}, to
transformer- and flow-based scaling \cite{peebles2023dit,esser2024sd3}.
Recent systems further unify generation, multimodal understanding, and
interactive editing. GPT-Image, GPT-Image-2, Nano Banana and Nano
Banana Pro, Qwen-Image, and OmniGen2 support capabilities such as
multi-image conditioning, conversational editing, text rendering, and
modification of real photographs
\cite{openai2025gptimage,openai2026gptimage2,
google2025nanobanana,google2025nanobananapro,
wu2025qwenimage,wu2025omnigen2}.
These capabilities make synthetic images increasingly realistic,
controllable, and easy to revise, increasing authenticity risks for
image verification.

\paragraph{AI-generated image detection.}
AI-generated image detection has evolved from recognizing
generator-specific artifacts to learning more general and semantically
informed notions of authenticity. Early methods exploit GAN fingerprints,
upsampling artifacts, global texture inconsistency, and local frequency
or texture irregularities
\cite{wang2020cnn,liu2020global}; these cues can be highly effective on
represented generators, but often weaken under unseen architectures,
compression, resizing, and editing. Later methods improve
cross-generator robustness by transferring pretrained visual
representations~\cite{ojha2023universal}, using reconstruction or
diffusion-based discrepancies~\cite{wang2023dire}, or expanding training
diversity across many generators, as in Community
Forensics~\cite{park2024community}. Recent detectors further introduce
heterogeneous forensic priors: SDAIE learns a photographic prior from
camera metadata using only real camera images~\cite{zhong2025sdaie};
AIDE combines semantic representations with high- and low-frequency
evidence~\cite{yan2025sanity}; SIDA unifies classification
and textual explanation in a multimodal model~\cite{huang2024sida}; and
Veritas incorporates planning and self-reflection for pattern-aware
reasoning~\cite{tan2025veritas}. These methods improve generalization,
but their strengths remain generator- and evidence-dependent: a cue that
is reliable for one generator family may be weak or misleading for
another. FACT therefore does not introduce another fixed forensic prior.
Instead, it treats existing priors as frozen callable experts and learns
an image-conditioned policy for deciding which evidence should be
acquired before making a decision.

\paragraph{Self-improving agents and agentic forensics.}
Tool-using agents interleave reasoning with external actions, and recent
self-improving agents learn from reflection, interaction feedback,
trajectories, or reusable skills
\cite{yao2023react,shinn2023reflexion,madaan2023selfrefine,
zhao2024expel,zheng2025skillweaver,chen2025step,yang2026skillopt}.
In image forensics, UniShield is the closest published agentic
framework: it predicts a broad forgery category, selects a corresponding
detector, and generates a report \cite{huang2025unishield}. FACT differs
by learning a full tool-use investigation procedure rather than a
category-to-detector routing rule: it compiles an execution-verified
forensic skill into real action--observation trajectories, distills them
into a compact agent, and refines the policy for cost-aware stopping.

\section{Method}
\label{sec:method}

\subsection{Overview and Problem Formulation}
\label{sec:overview}

Given an image $x$, FACT aims to determine whether it is real or
AI-generated. To support this decision, FACT first constructs a forensic
toolbox $\mathcal{U}$ from two complementary evidence sources: frozen
model-based experts and lightweight signal-level probes. The goal is to
learn a policy $\pi_\theta$ that selectively uses this toolbox before
predicting $\hat{y}\in\{\mathrm{real},\mathrm{AI}\}$. At step $t$, the
policy observes the interaction history
\[
h_t=(x,a_1,o_1,\ldots,a_{t-1},o_{t-1}),
\]
and either calls a tool $a_t\in\mathcal{U}$ to obtain observation $o_t$,
or stops and outputs a final verdict. The complete trajectory and its
acquisition cost are
\[
\tau=(x,a_1,o_1,\ldots,a_T,o_T,\hat{y}), \qquad
C(\tau)=\sum_{t:a_t\in\mathcal{U}} c(a_t),
\]
where $c(a_t)$ is the normalized cost of tool $a_t$.

\begin{figure*}[t]
    \centering
    \IfFileExists{figures/fact_framework.pdf}{
        \includegraphics[width=0.96\textwidth]{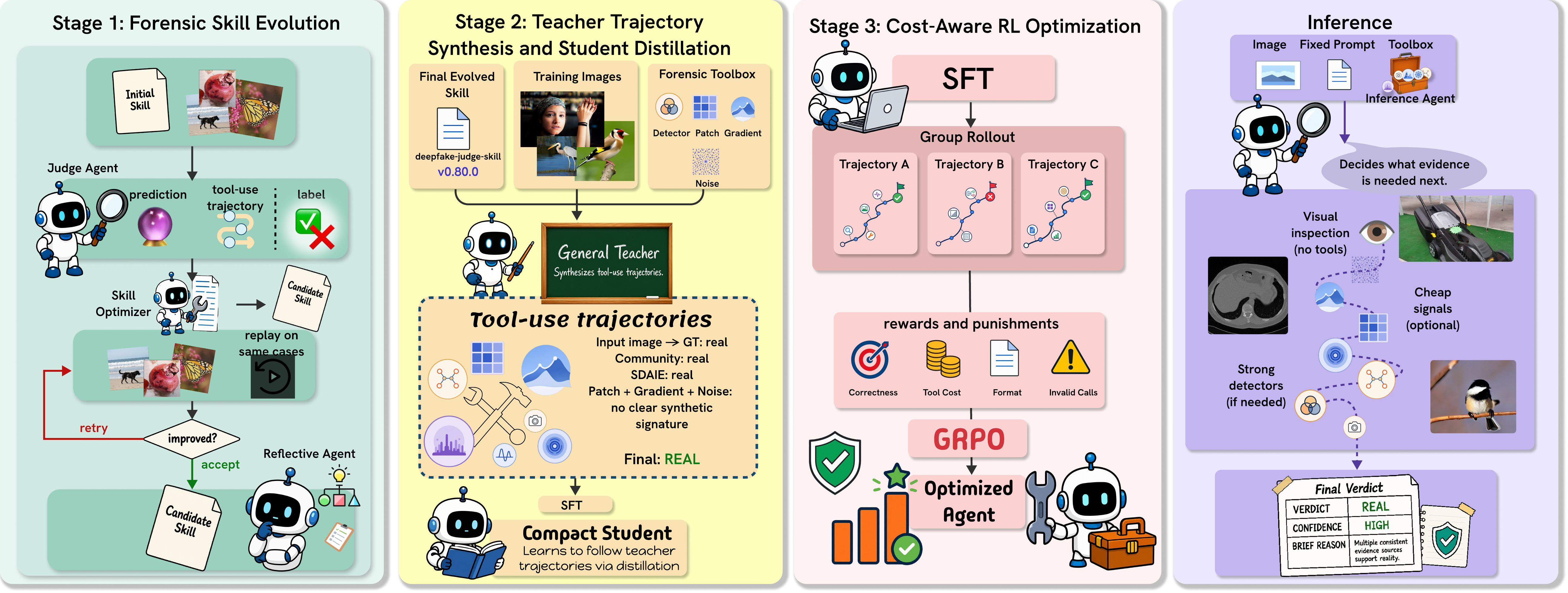}
    }{
        \IfFileExists{figures/fact_framework.png}{
            \includegraphics[width=0.96\textwidth]{figures/fact_framework.png}
        }{
            \fbox{\parbox{0.94\textwidth}{
            \centering
            \textbf{Evolve:} executions $\rightarrow$ verified skill
            \qquad $\Longrightarrow$ \qquad
            \textbf{Distill:} skill $\rightarrow$ trajectories
            $\rightarrow$ compact student
            \qquad $\Longrightarrow$ \qquad
            \textbf{Refine:} grouped rollouts $\rightarrow$
            cost-aware policy
            }}
        }
    }
  \caption{
Overview of FACT.
FACT trains a forensic agent with an Evolve--Distill--Refine pipeline.
Skill evolution learns an executable investigation procedure from judged
cases; trajectory distillation converts the skill into supervised
action--observation tool-use data; and cost-aware GRPO teaches the agent
when additional evidence is worth its cost. At inference, the trained
agent adaptively calls forensic tools or stops with a final real/AI
decision.
}
    \label{fig:framework}
\end{figure*}

As shown in Fig.~\ref{fig:framework}, FACT learns $\pi_\theta$ in three
stages. \emph{Stage I} uses a general AI model to analyze labeled
images with the toolbox and iteratively improve a textual forensic skill
$S$; candidate skill revisions are accepted only after rerunning the
tool-use process and verifying that they improve accuracy. \emph{Stage
II} fixes the verified skill $S^\star$ and uses it to generate real
action--observation trajectories in the same tool environment, which are
then distilled into a compact policy by SFT. \emph{Stage III}
refines the distilled policy with cost-aware GRPO, rewarding correct
final decisions while penalizing tool cost, malformed answers, and
invalid calls. During inference, FACT uses only the trained policy and
the toolbox: for each image, the policy decides whether to answer
directly, call more evidence tools, or stop.

\subsection{Forensic Toolbox Construction}
\label{sec:toolbox}

Before training the agent, we first examine how existing detectors behave
on different generator families. As shown in
Table~\ref{tab:detector_specialization}, detectors that are strong on
one generator can be weaker on another. This observation suggests that the goal should not be to
choose a single detector as the universal solution. Instead, FACT
organizes complementary detectors and signal analyses into a toolbox
$\mathcal{U}$, so that the agent can learn which evidence is useful for
each image.

\paragraph{Model-based experts.}
The first part of the toolbox consists of five frozen model-based
experts. These experts capture different types of forensic evidence and provide
complementary signals for AI-generated image detection. 
\textbf{SDAIE} models the photographic regularities of real camera
images through EXIF-induced supervision~\cite{zhong2025sdaie}.
\textbf{Community Forensics} emphasizes broad discriminative coverage by
training across thousands of generators~\cite{park2024community}.
\textbf{AIDE} combines semantic features with high- and low-frequency
evidence~\cite{yan2025sanity}. \textbf{SIDA} brings multimodal reasoning and textual explanation into the detection
process~\cite{huang2024sida}. \textbf{Veritas} uses pattern-aware
planning and self-reflection to reason about synthetic
artifacts~\cite{tan2025veritas}. These experts therefore capture
different cues, including camera-photo consistency, large-scale
generator diversity, semantic plausibility, multimodal reasoning, and
reasoning-based artifact analysis. Their strengths across
generator subsets motivate using them as callable experts rather than
choosing one as the fixed detector.

\paragraph{Signal-level probes.}
The second part of the toolbox contains eight lightweight signal-level
probes. These probes are included because low-level texture, frequency,
noise, and compression statistics have long been useful for fake-image
detection~\cite{wang2020cnn,liu2020global}, and they provide evidence
that is different from the final prediction of a neural detector. FFT
and phase statistics describe spectral behavior; wavelet and Laplacian
analyses measure multiscale texture; noise residuals and JPEG analysis
capture residual and compression consistency; gradient statistics and
local patch consistency describe local structural regularity. These
probes do not replace model-based experts. Rather, they give the agent
simple and interpretable measurements that can be combined with expert
outputs when deciding whether an image is real or AI-generated.

\subsection{Stage I: Tool-Use Skill Learning}
\label{sec:skill_evolution}

With the toolbox $\mathcal{U}$ fixed, Stage I learns how a 
general model $M_G$ should use it for forensic judgment. The goal is not
only to decide whether an image is real or AI-generated, but also to
learn a reusable investigation procedure: which tools to call, why they
are useful, how to interpret their outputs, how to handle conflicting
evidence, and when to stop. Following recent agent works that store
reusable experience as skills or reflection-derived procedures
\cite{zhao2024expel,zheng2025skillweaver,yang2026skillopt}, FACT
represents this procedure as a textual forensic skill $S$. In our
implementation, $S$ is a structured instruction covering tool selection,
evidence interpretation, conflict handling, stopping criteria, and
final-answer format. Updating $S$ lets $M_G$ become more specialized for
AI-generated image detection without finetuning its parameters.

At update step $r$, FACT samples a labeled skill-update batch
$\mathcal{B}_r=\{(x_i,y_i)\}_{i=1}^{m}$. The current skill $S_r$ is
given to $M_G$, which analyzes each image with access to the toolbox
$\mathcal{U}$. During the investigation, $M_G$ sees the image and the
tool observations it requests, but not the ground-truth label. Each run
is recorded as a trajectory
\[
\tau_i=(x_i,a_{i,1},o_{i,1},\ldots,a_{i,T_i},o_{i,T_i},\hat y_i),
\]
where $\hat y_i$ is the final prediction. After the batch is judged, the
labels are revealed and the errors are summarized as
\[
\mathcal{D}_r=\{(x_i,y_i,\hat y_i,\tau_i)\}_{i=1}^{m}.
\]
The model then uses the current skill and this error summary to propose
a revised skill:
\[
\widetilde S_r=\mathrm{UpdateSkill}_{M_G}(S_r,\mathcal{D}_r).
\]

FACT evaluates the proposed revision on a separately sampled labeled
verification batch
\[
\mathcal{V}_r=\{(x_j^{(v)},y_j^{(v)})\}_{j=1}^{n},
\]
which is not used to construct $\mathcal{D}_r$ or propose
$\widetilde S_r$. Let
\[
(\hat y_j^{(v)}(S),\tau_j^{(v)}(S))
=\mathcal{F}_{M_G}(S,x_j^{(v)};\mathcal{U})
\]
denote running $M_G$ on verification image $x_j^{(v)}$ with skill $S$
and toolbox $\mathcal{U}$. The verification accuracy is
\[
\mathcal{J}(S;\mathcal{V}_r)
=
\frac{1}{|\mathcal{V}_r|}
\sum_{(x_j^{(v)},y_j^{(v)})\in\mathcal{V}_r}
\mathbf{1}\!\left[\hat y_j^{(v)}(S)=y_j^{(v)}\right].
\]
The skill is updated by the following replay check:
\[
S_{r+1}=
\begin{cases}
\widetilde S_r,
& \mathcal{J}(\widetilde S_r;\mathcal{V}_r)>
  \mathcal{J}(S_r;\mathcal{V}_r),\\
S_r, & \text{otherwise}.
\end{cases}
\]
Thus, a revised instruction is accepted only when it improves rerun
tool-use performance on the separate verification batch.

Beyond single-batch updates, FACT also summarizes recent error summaries
and proposes broader skill revisions. For example, if the model
repeatedly over-trusts frequency artifacts on compressed real images,
the skill can be revised to check compression consistency before making
a final decision. If the model repeatedly overlooks spatially inconsistent evidence, the
skill can place greater emphasis on patch-level consistency checks. These broader
updates are retained only after the same replay check. The output of
Stage I is a verified forensic skill $S^\star$, which provides the
procedure used to generate training trajectories in Stage II.
Representative skill revisions and the complete initial and evolved
skill texts are provided in the supplementary material.

\subsection{Stage II: Distilling Tool-Use Trajectories}
\label{sec:distillation}

Stage I produces a verified forensic skill $S^\star$, but this skill is
still a textual procedure used by a general model. For deployment,
we need a compact model that can execute the procedure directly: it
should know when to call a tool, how to react to the returned evidence,
and when to stop. Stage II therefore converts the skill into supervised
tool-use trajectories and distills them into a compact policy
$\pi_\theta$.

For each labeled training image $(x_i,y_i)$, the general model $M_G$
follows $S^\star$ and interacts with the toolbox $\mathcal{U}$. It
selects a tool, receives the tool output, and then decides the next
action until it produces a final decision. We record this process as
\[
\tau_i=(x_i,a_{i,1},o_{i,1},\ldots,
a_{i,T_i},o_{i,T_i},y_i),
\]
where $a_{i,t}$ is a tool call or a stopping action, $o_{i,t}$ is the
corresponding tool output, and $y_i$ provides the supervised final
answer. The key supervision is not only the final label, but the
intermediate action sequence showing how evidence changes the next
decision.

We serialize each trajectory into a training sequence containing the
image, task instruction, tool definitions, previous tool calls, returned
tool outputs, and the next action to imitate. The compact policy is
trained by supervised fine-tuning:
\[
\mathcal{L}_{\mathrm{SFT}}
= -\frac{1}{\sum_{i,t}m_{i,t}}
\sum_{i,t}m_{i,t}
\log \pi_\theta(z_{i,t}\mid x_i,z_{i,<t}),
\]
where $z_{i,t}$ is a token in the serialized trajectory and
$m_{i,t}=1$ only for tokens that the agent should generate, such as tool
calls, stopping decisions, and the final answer. Tokens from tool
outputs are used as context but are not prediction targets.

After this stage, $\pi_\theta$ becomes a compact forensic agent trained
to imitate the tool-use behavior induced by $S^\star$. It has learned
the basic investigation procedure, while Stage III further optimizes how
much evidence should be acquired under a tool-cost budget.

\subsection{Stage III: Cost-Aware Tool-Use Refinement}
\label{sec:grpo}

The SFT policy has learned how to conduct a forensic investigation, but
supervised imitation does not directly optimize the deployment objective:
making a correct decision under a limited evidence budget. Stage III
therefore refines the distilled policy with cost-aware GRPO. 

For each training image $x_i$, we sample a group of $K$ complete
trajectories
\[
\{\tau_i^k\}_{k=1}^{K}, \qquad
\tau_i^k=(x_i,a_{i,1}^k,o_{i,1}^k,\ldots,a_{i,T_i^k}^k,
o_{i,T_i^k}^k,\hat y_i^k),
\]
from the current policy. Each trajectory is scored by a task--cost
reward:
\[
R(\tau_i^k,y_i)
=
R_{\mathrm{task}}(\hat y_i^k,y_i)
-\lambda C(\tau_i^k)
-R_{\mathrm{format}}(\tau_i^k)
-R_{\mathrm{invalid}}(\tau_i^k).
\]
The task term rewards the final decision,
\[
R_{\mathrm{task}}(\hat y,y)=
\begin{cases}
+1, & \hat y=y,\\
-1, & \hat y\neq y,\\
0, & \hat y=\mathrm{uncertain},\\
-1.2, & \text{no valid final answer},
\end{cases}
\]
while the cost term penalizes cumulative tool acquisition,
\[
C(\tau)=\sum_{t:a_t\in\mathcal{U}} c(a_t).
\]
The remaining terms penalize interaction failures:
\[
\begin{aligned}
R_{\mathrm{format}}(\tau)
&=0.2\,\mathbf{1}[\text{malformed final response}],\\
R_{\mathrm{invalid}}(\tau)
&=0.2\,N_{\mathrm{invalid}}(\tau),
\end{aligned}
\]
where $N_{\mathrm{invalid}}(\tau)$ is the number of invalid tool calls.

GRPO compares trajectories sampled for the same image. For trajectory
$\tau_i^k$, its group-relative advantage is
\[
A_i^k=
\frac{R(\tau_i^k,y_i)-\frac{1}{K}\sum_{\ell=1}^{K}R(\tau_i^\ell,y_i)}
{\mathrm{std}_{\ell}(R(\tau_i^\ell,y_i))+\epsilon}.
\]
Let $z_{i,k,t}$ denote an agent-generated token in trajectory
$\tau_i^k$, including tool calls and final-answer tokens, and let
$m_{i,k,t}$ mask these trainable tokens. The clipped GRPO objective is
\[
g_{i,k,t}
=
\min\!\left(
\rho_{i,k,t} A_i^k,\,
\mathrm{clip}(\rho_{i,k,t},1-\epsilon,1+\epsilon) A_i^k
\right).
\]

\[
\mathcal{L}_{\mathrm{GRPO}}
=
-\frac{1}{\sum_{i,k,t}m_{i,k,t}}
\sum_{i,k,t} m_{i,k,t} g_{i,k,t}
+\beta D_{\mathrm{KL}}(\pi_\theta\|\pi_{\mathrm{ref}}).
\]
where
\[
\rho_{i,k,t}
=
\frac{\pi_\theta(z_{i,k,t}\mid x_i,z_{i,k,<t})}
{\pi_{\mathrm{old}}(z_{i,k,t}\mid x_i,z_{i,k,<t})},
\]
and $\pi_{\mathrm{ref}}$ is the fixed SFT reference policy.

This refinement does not ask RL to discover tool use from scratch.
Instead, because the policy already imitates valid trajectories from
Stage II, the reward mainly teaches whether another tool call improves
the final decision enough to justify its cost. The coefficient
$\lambda$ controls this evidence budget: increasing $\lambda$ encourages
the agent to avoid unnecessary calls, while the task reward prevents it
from stopping too early.

\paragraph{Inference.}
At inference time, FACT uses the trained forensic agent $\pi_\theta$
to judge each test image. The agent predicts a tool-use sequence over the
forensic toolbox $\mathcal{U}$, where each step either calls a tool or
terminates with \textsc{Stop}. Returned observations are added to the
history and used to choose the next action. Once the agent stops, it
outputs a real/AI verdict and brief rationale based
on the evidence it acquired. The skill-learning and trajectory-synthesis
components are used only during training.

\section{Experiments}
\label{sec:experiments}

\subsection{Experimental Setup}

\paragraph{Benchmarks.}
We evaluate FACT on two balanced internal benchmarks and four public
benchmarks. The internal benchmarks are designed to isolate two
deployment scenarios that are not fully covered by existing test sets.
\textbf{STD3K} contains 3,200 images, with 1,600 generated images from
21 generator/source groups and 1,600 real images from LIVE, CLIVE,
Flickr8K, and TID2013
\cite{sheikh2006live,ghadiyaram2016clive,hodosh2013flickr8k,
ponomarenko2015tid2013}. It measures robustness under broad source
diversity rather than performance on a single generator family.
\textbf{NewGen-900} contains 900 images, with 450 generated images and
450 real images from the same real-image sources. The fake images are
drawn from nine recent generators: Nano Banana, Nano Banana Pro,
GPT-Image-1, GPT-Image-2, SD3.5-Large, SD3-Medium, OmniGen2,
Qwen-Image, and Z-Image Turbo. These generator families are held out from all FACT training stages,
so NewGen-900 directly evaluates open-world cross-generator
generalization.

For public evaluation, we use benchmarks that emphasize different
failure modes. \textbf{Chameleon} contains visually challenging
AI-generated images that are often misclassified as real by existing
detectors~\cite{yan2025sanity}; we evaluate on its official split. \textbf{LOKI} evaluates synthetic-data
detection under a multimodal LMM protocol with real/synthetic judgments
and explanation-oriented questions~\cite{ye2025loki}; we report balanced
accuracy on its image-detection subset to
match the public protocol. \textbf{AIGCDetectionBenchmark} covers a broad set of AI image
generators, including 17 prevalent generative
models~\cite{zhong2023patchcraft}; we use its official image-detection
test split, which also enables direct comparison with UniShield.
\textbf{HydraFake} extends the evaluation beyond full-image generation
to broader image-forgery settings, including face swapping,
reenactment, attribute editing, personalization, relighting,
restoration, and cross-domain deepfakes~\cite{tan2025veritas}.
We report the average accuracy across its official evaluation splits~\cite{tan2025veritas}.
Detailed dataset composition, sampling rules, and benchmark documentation
are provided in the supplementary material.

\paragraph{Baselines and metrics.}
We compare FACT with the five frozen model-based experts in the
toolbox---SDAIE, Community Forensics, AIDE, SIDA, and Veritas---and
their unweighted majority vote. For the public benchmarks, we
additionally include published results under the corresponding benchmark
protocols. Detailed baseline descriptions and the exact provenance of
externally reported results are provided in the supplementary material.

We report accuracy on STD3K, NewGen-900, Chameleon, and
AIGCDetectionBenchmark; balanced accuracy on LOKI,
$\mathrm{BAcc}=\tfrac{1}{2}(\mathrm{TPR}+\mathrm{TNR})$; and average
accuracy across the official HydraFake evaluation splits. Efficiency is measured by the
average normalized acquisition cost $C(\tau)$.

\paragraph{Implementation details.}
FACT uses GPT-5.1 for skill evolution and trajectory synthesis, and
Qwen3-VL-8B-Instruct~\cite{bai2025qwen3vl} as the compact forensic
agent. Full training configurations, optimization settings, and
tool-cost calibration are provided in the supplementary material.


\subsection{Comparison with Existing Works}

Tables~\ref{tab:internal_results}--\ref{tab:hydrafake_results} compare
FACT with its constituent experts, majority voting, and representative
published baselines on the internal and public benchmarks. Complete
baseline results and their provenance are provided in the supplementary
material. Unless otherwise stated, FACT denotes the Stage III refined
checkpoint.

\begin{table}[!t]
\centering
\small
\setlength{\tabcolsep}{6.0pt}

\begin{tabular}{lrr}
\toprule
Method & STD3K & NewGen-900 \\
\midrule
Community~\cite{park2024community}
    & 78.56 & 72.67 \\
Veritas~\cite{tan2025veritas}
    & 40.78 & 43.11 \\
SIDA~\cite{huang2024sida}
    & 77.38 & 61.00 \\
SDAIE~\cite{zhong2025sdaie}
    & 59.53 & 76.89 \\
AIDE~\cite{yan2025sanity}
    & 44.31 & 73.44 \\
Majority Vote
    & 72.97 & 75.33 \\
\midrule
FACT
    & \textbf{91.40} & \textbf{82.44} \\
\bottomrule
\end{tabular}

\caption{
Comparison on the internal benchmarks.
All values are accuracies in percentages. All methods are evaluated
using our fixed evaluation pipeline.
}
\label{tab:internal_results}
\end{table}

On \textbf{STD3K}, the final refined FACT checkpoint reaches 91.40\%,
exceeding the strongest constituent expert by 12.84 percentage points
and demonstrating robustness across diverse generator and real-image
sources.

On \textbf{NewGen-900}, the final refined FACT checkpoint obtains
82.44\%, outperforming the strongest constituent expert by 5.55 points.
All nine generator families are held out from skill evolution,
trajectory distillation, and RL refinement, demonstrating open-world
generalization without generator-specific fine-tuning.

On \textbf{Chameleon}, FACT achieves 87.53\%, exceeding DefakerOne,
the strongest previously reported comparison, by 2.83 points. This
indicates strong performance on visually challenging synthetic images.

\begin{table}[t]
\centering
\small
\setlength{\tabcolsep}{5.0pt}

\begin{tabular}{@{}lr@{}}
\toprule
Method & Accuracy \\
\midrule
NPR$^{\dagger}$~\cite{tan2024rethinking}          & 57.81 \\
AIDE$^{\dagger}$~\cite{yan2025sanity}             & 65.77 \\
Community~\cite{park2024community}                 & 78.44 \\
Veritas~\cite{tan2025veritas}                      & 58.23 \\
SIDA~\cite{huang2024sida}                          & 66.50 \\
SDAIE~\cite{zhong2025sdaie}                       & 58.86 \\
Majority Vote                                      & 65.97 \\
Ivy-xDetector$^{\ddagger}$~\cite{ivyfake}          & 73.17 \\
DefakerOne$^{\ddagger}$~\cite{defakerone}          & \underline{84.70} \\
\midrule
FACT                                                & \textbf{87.53} \\
\bottomrule
\end{tabular}

\caption{
Comparison on Chameleon.
$\dagger$: Table~5 of \cite{yan2025sanity};
$\ddagger$: cited method papers. Unmarked results use our fixed
pipeline.
}
\label{tab:chameleon_results}
\end{table}

On \textbf{LOKI}, FACT achieves 88.35\% balanced accuracy, surpassing
EvoGuard by 1.97 percentage points under the reported public protocol.
This shows that the learned policy remains effective under the
multimodal evaluation setting.

On \textbf{AIGCDetectionBenchmark}, FACT reaches 98.77\%, exceeding the
strongest comparison by 1.97 points. The gain over individual detectors
and majority voting supports image-conditioned evidence acquisition over
static aggregation.

On \textbf{HydraFake}, FACT achieves 94.44\% average accuracy,
outperforming Veritas by 3.74 points and showing that FACT remains effective beyond fully generated images,
including on deepfakes and edited images. Across all six
benchmarks, FACT outperforms every constituent expert, majority voting,
and the strongest available published comparison.

\begin{table}[t]
\centering
\small
\setlength{\tabcolsep}{5.0pt}

\begin{tabular}{@{}lr@{}}
\toprule
Method & BAcc. \\
\midrule
Effort$^{\dagger}$~\cite{yan2025effort}                & 74.09 \\
FakeVLM$^{\dagger}$~\cite{wen2025fakevlm}              & 81.62 \\
MIRROR$^{\dagger}$~\cite{liu2026mirror}                & 85.23 \\
AIDE$^{\dagger}$~\cite{yan2025sanity}                  & 73.70 \\
FakeShield$^{\dagger}$~\cite{xu2025fakeshield}         & 61.55 \\
SIDA$^{\dagger}$~\cite{huang2024sida}                  & 61.92 \\
FakeReasoning$^{\dagger}$~\cite{gao2025fakereasoning}  & 64.78 \\
Forensic-MoE$^{\dagger}$~\cite{fang2025forensicmoe}    & 74.92 \\
EvoGuard$^{\dagger}$~\cite{zhu2026evoguard}            & \underline{86.38} \\
Community~\cite{park2024community}                      & 82.80 \\
Veritas~\cite{tan2025veritas}                           & 65.95 \\
SDAIE~\cite{zhong2025sdaie}                            & 82.85 \\
Majority Vote                                           & 83.20 \\
\midrule
FACT                                                     & \textbf{88.35} \\
\bottomrule
\end{tabular}

\caption{
Comparison on LOKI.
$\dagger$: Table~2 of \cite{zhu2026evoguard}. Unmarked results use our
fixed pipeline.
}
\label{tab:loki_results}
\end{table}

\begin{table}[t]
\centering
\small
\setlength{\tabcolsep}{5.0pt}

\begin{tabular}{@{}lr@{}}
\toprule
Method & Accuracy \\
\midrule
AIDE$^{\dagger}$~\cite{yan2025sanity}              & 92.80 \\
FakeVLM$^{\dagger}$~\cite{wen2025fakevlm}          & 81.00 \\
UniShield$^{\dagger}$~\cite{huang2025unishield}    & 94.20 \\
Community~\cite{park2024community}                  & 96.80 \\
Veritas~\cite{tan2025veritas}                       & 69.60 \\
SIDA~\cite{huang2024sida}                           & 59.90 \\
SDAIE~\cite{zhong2025sdaie}                        & 96.00 \\
ReAlign$^{\ddagger}$~\cite{realign}                 & 96.14 \\
Majority Vote                                       & 95.50 \\
\midrule
FACT                                                 & \textbf{98.77} \\
\bottomrule
\end{tabular}

\caption{
Comparison on AIGCDetectionBenchmark.
$\dagger$: Table~6 of \cite{huang2025unishield};
$\ddagger$: cited method paper. Unmarked results use our fixed pipeline.
}
\label{tab:aigcdet_results}
\end{table}

\begin{table}[t]
\centering
\small
\setlength{\tabcolsep}{1.4pt}

\begin{tabular}{@{}c@{\hspace{4pt}}c@{}}
\begin{tabular}[t]{@{}lr@{}}
\multicolumn{2}{c}{(a)} \\
\toprule
Method & Acc. \\
\midrule
FreqNet$^{\dagger}$         & 64.60 \\
ProDet$^{\dagger}$          & 80.60 \\
NPR$^{\dagger}$             & 69.80 \\
AIDE$^{\dagger}$            & 70.60 \\
Co-SPY$^{\dagger}$          & 84.70 \\
D$^{3\dagger}$              & 81.10 \\
Effort$^{\dagger}$          & 82.20 \\
Qwen2.5-VL-7B$^{\dagger}$   & 54.10 \\
InternVL3-8B$^{\dagger}$    & 58.30 \\
MiMo-VL-7B$^{\dagger}$      & 72.50 \\
GLM-4.1V-9B-T$^{\dagger}$   & 61.70 \\
GPT-4o$^{\dagger}$          & 60.80 \\
\bottomrule
\end{tabular}
&
\begin{tabular}[t]{@{}lr@{}}
\multicolumn{2}{c}{(b)} \\
\toprule
Method & Acc. \\
\midrule
Gemini-2.5-Pro$^{\dagger}$  & 78.90 \\
M2F2-Det$^{\dagger}$        & 63.20 \\
FakeShield$^{\dagger}$      & 60.80 \\
SIDA-7B$^{\dagger}$         & 76.30 \\
SIDA-13B$^{\dagger}$        & 69.80 \\
FFAA$^{\dagger}$            & 64.00 \\
FakeVLM$^{\dagger}$         & 77.30 \\
Veritas$^{\dagger}$         & \underline{90.70} \\
AIDE (ours)                 & 61.60 \\
SDAIE (ours)                & 68.30 \\
Community (ours)            & 76.30 \\
FACT                         & \textbf{94.44} \\
\bottomrule
\end{tabular}
\end{tabular}

\caption{Comparison on HydraFake. Values are average accuracies in
percentages. $\dagger$: Table~1 of \cite{tan2025veritas}; unmarked
results use our fixed pipeline. ``T'' abbreviates Think.}
\label{tab:hydrafake_results}
\end{table}

The supplementary material additionally provides qualitative case
analyses, full agent conversation traces, and the human evaluation.

\subsection{Ablation Studies}

\begin{table}[t]
\centering
\small
\setlength{\tabcolsep}{1.1pt}

\begin{tabular}{@{}c@{\hspace{4pt}}c@{}}
\begin{tabular}[t]{@{}lcccrr@{}}
\multicolumn{6}{c}{(a) Training stages} \\
\toprule
Var. & S & D & R & STD & New \\
\midrule
Direct   & -- & -- & -- & 87.90 & 50.80 \\
Skill    & \checkmark & -- & -- & 89.40 & 75.10 \\
Distill. & \checkmark & \checkmark & -- &
           \textbf{91.53} & 81.89 \\
Refine   & \checkmark & \checkmark & \checkmark &
           91.40 & \textbf{82.44} \\
\bottomrule
\end{tabular}
&
\begin{tabular}[t]{@{}ccr@{}}
\multicolumn{3}{c}{(b) Toolbox} \\
\toprule
Sig. & Exp. & Acc. \\
\midrule
\checkmark & --          & 82.90 \\
--         & \checkmark  & 82.50 \\
\checkmark & \checkmark  & \textbf{90.70} \\
\bottomrule
\end{tabular}
\end{tabular}

\caption{Pipeline and toolbox ablations. In (a), S, D, and R denote
skill learning, trajectory distillation, and GRPO refinement; the
variants are Direct, Skill-guided, Distilled, and Refined. In (b), Sig.
and Exp. denote signal probes and model-based experts.}
\label{tab:stage_tool_ablation}
\end{table}

\begin{table}[!ht]
\centering
\small
\setlength{\tabcolsep}{2.2pt}
\begin{tabular}{ccccrr}
\toprule
Task & Fmt. & Valid & Cost & Acc. & Acq. \\
\midrule
\checkmark & --         & --         & --         &
81.78 & 1.5300 \\
\checkmark & \checkmark & --         & --         &
82.00 & 1.5700 \\
\checkmark & \checkmark & \checkmark & --         &
\textbf{82.56} & 1.5500 \\
\checkmark & \checkmark & \checkmark & \checkmark &
82.44 & \textbf{1.3835} \\
\bottomrule
\end{tabular}
\caption{Reward-component ablation on NewGen-900. Reward terms are
added cumulatively; the full reward uses $\lambda=0.01$.}
\label{tab:reward_cost}
\end{table}

\paragraph{Contribution of each training stage.}
Table~\ref{tab:stage_tool_ablation}(a) evaluates skill learning,
trajectory distillation, and cost-aware refinement. Directly querying
GPT-5.1 reaches 87.90\% on STD3K but only 50.80\% on NewGen-900.
Adding the learned skill raises performance to 89.40\% and 75.10\%,
showing that Stage I contributes procedural knowledge beyond direct
visual judgment.

Distilling skill-guided action--observation trajectories further improves
performance to 91.53\% and 81.89\%. Cost-aware GRPO produces the final
FACT checkpoint, changing accuracy to 91.40\% on STD3K and 82.44\% on
NewGen-900. Thus, refinement trades a marginal 0.13-point decrease on
STD3K for a 0.55-point gain on NewGen-900. In the reward-component
experiment, the full cost-aware reward achieves 82.44\% accuracy while
reducing normalized acquisition cost from 1.5500 to 1.3835
(Table~\ref{tab:reward_cost}). 

\paragraph{Effect of toolbox composition.}
Table~\ref{tab:stage_tool_ablation}(b) compares signal probes,
model-based experts, and their combination on STD3K + NewGen-900. The
two families alone obtain 82.90\% and 82.50\%, while combining them
reaches 90.70\%. This confirms that low-level measurements and
high-level detector outputs provide complementary evidence.

\paragraph{Effect of reward components.}
Table~\ref{tab:reward_cost} shows that format and valid-tool penalties
improve NewGen-900 accuracy from 81.78\% to 82.56\%. Adding tool cost
yields 82.44\% accuracy and lowers acquisition cost from 1.5500 to
1.3835.

\section{Conclusion}

We presented FACT, a forensic agent that reframes AI-generated image
detection as learning an investigation procedure rather than training
another fixed detector. FACT converts image-level supervision into a
tool-use skill, compiles it into action--observation trajectories, and
distills them into a cost-aware agent. Across benchmarks, FACT improves
over individual experts and static voting while generalizing to recent
generators held out from all FACT training stages.

The results show that evidence acquisition and stopping should be
optimized jointly. The refined policy preserves detection performance
while reducing normalized acquisition cost, avoiding redundant calls
without abandoning difficult cases.

FACT suggests that robustness may depend less on a universal detector
and more on learning which evidence each image requires. Additional
qualitative analyses and limitations are provided in the supplementary
material. Upon publication, we will release the code, trained model, and
benchmark data and metadata, subject to source licenses.


\bibliography{references}

\end{document}